# Quantum Low Entropy based Associative Reasoning – QLEAR Learning

**Marko V. Jankovic,**

Department of Emergency Medicine, Bern University Hospital "Inselspital", Bern, Switzerland and ARTORG Center for Biomedical Engineering Research, University of Bern, Switzerland

**Abstract** – In this paper, we propose the classification method based on a learning paradigm we are going to call Quantum Low Entropy based Associative Reasoning (QLEAR learning). The approach is based on the idea that classification can be understood as supervised clustering, where a quantum entropy in the context of the quantum probabilistic model, will be used as a "capturer" (measure, or external index) of the "natural structure" of the data. By using quantum entropy we don't make any assumption about linear separability of the data that are going to be classified. The basic idea is to find close neighbours to a query sample and then use relative change in the quantum entropy as a measure of similarity of the newly arrived sample with the representatives of interest. In other words, method is based on calculation of quantum entropy of the referent system and its relative change with the addition of the newly arrived sample. Referent system consists of vectors that represent individual classes and that are the most similar, in Euclidean distance sense, to the vector that is analyzed. Here, we analyse the classification problem in the context of measuring similarities to prototype examples of categories. While nearest neighbour classifiers are natural in this setting, they suffer from the problem of high variance (in bias-variance decomposition) in the case of limited sampling. Alternatively, one could use machine learning techniques (like support vector machines) but they involve time-consuming optimization. Here we propose a hybrid of nearest neighbour and machine learning technique which deals naturally with the multiclass setting, has reasonable computational complexity both in training and at run time, and yields excellent results in practice.

## I. Introduction

It is well known that the field of pattern recognition is concerned with the automatic discovery of regularities in data through the use of computer algorithms and with the use of these regularities to take actions such as classifying the data into different categories. There are many algorithms based on different theoretical backgrounds that could be used for pattern recognition in practical applications (see e.g. [1]). Generally, most of the algorithms are applied in areas like classification, regression or change point detection.

Recently, it has been shown that a probabilistic model based on two of the main concepts in quantum physics – a density matrix and the Born rule, can be suitable for the modeling of learning algorithms in biologically plausible artificial neural networks framework. It has been shown that the proposed probabilistic interpretation is suitable for modeling on-line learning algorithms for Independent /Principal/Minor Component Analysis [2-5], which could be realized on parallel hardware based on very simple computational units. Also, it has been shown that the quantum entropy of the system, related to that model, can be successfully used in the problems like change point or anomalies detection [6, 7] as well as simple classification problems [7].

Here another application of the proposed quantum probabilistic model is going to be presented. A general paradigm called QLEAR learning (Quantum Low Entropy based Associative Reasoning) would be presented and tested in classification context. Proposed method potentially can overcome the problem that classifier performance depends greatly on the characteristics of the data to be classified. It is known that until now, there is no single classifier that works best on all given problems (a phenomenon that may be explained by the no-free-lunch theorem). Here we will try to propose a classification algorithm that, actually, automatically adjusts its performance according to characteristics of the data on which it is applied. An interesting aspect is that proposed method inherently solves the problem of unbalanced classes (classes that have significantly different size). The proposed paradigm can be applied in any area in which standard classification techniques are applied.

We'll analyze only the case in which data is represented by vectors while generalization toward multiway data would not be discussed here. Also, some modification of the existing quantum probabilistic model, that have no ground in temporary quantum mechanics (at least currently), that were used to improve the quality of the model in the classification context, will not be discussed here.

## II.   Quantum Probability Model and Quantum Tsallis' Entropy

In quantum mechanics the transition from a deterministic description to a probabilistic one is done using a simple rule termed the Born rule. This rule states that the probability of an outcome (*a*) given a state (*Ψ*) is the square of their inner product ($a^T\Psi)^2$. This section is based on a similar section in [8, 3].

In quantum mechanics the Born rule is usually taken as one of the axioms. However, this rule has well established foundations. Gleason's theorem [9] states that the Born rule is the only consistent probability distribution for a Hilbert space structure. Wooters [10] has shown that by using the Born rule as a probability rule, the natural Euclidean metrics on a Hilbert space coincides with a natural notion of a statistical distance. Short review for some other justifications of the Born rule can be seen in [8].

The quantum probability model takes place in a Hilbert space H of finite or infinite dimension. A state is represented by a positive semidefinite linear mapping (a matrix ρ) from this space to itself, with a trace of 1, i.e. $\forall \Psi \in H$  $\Psi^T \rho \Psi \geq 0$, Tr(*ρ*) =1.  Such *ρ* is self adjoint and is called a density matrix.

Since *ρ* is self adjoint, its eigenvectors *Φ$_i$* are orthonormal and since it is positive semidefinite its eigenvalues $p_i$ are real and nonnegative $p_i \geq 0$. The trace of a matrix is equal to the sum of its eigenvalues, therefore $\sum_i p_i =1$.

The equality  $\rho = \sum_i p_i \Phi_i \Phi_i^T$  is interpreted as "the system is in state *Φ$_i$* with probability $p_i$". The state *ρ* is called the pure state if $\exists i$ s.t. $p_i = 1$. In this case, $\rho = \Psi\Psi^T$ for some normalized state vector *Ψ*, and the system is said to be in state *Ψ*.

A measurement M with an outcome *x* in some set *X* is represented by a collection of positive definite matrices $\{m_z\}_{z \in Z}$ such that $\sum_{z \in Z} m_z = \mathbf{1}$ (**1** being the identity matrix in H). Applying measurement M to state *ρ* produces an outcome *x* with probability

$$p_x(\rho) = \text{trace}(\rho m_x) \qquad (1)$$

This is the Born rule. Most quantum models deal with a more restrictive type of measurement called the von Neumann measurement, which involves a set of projection operators $m_a = aa^T$ for which $a^T a' = \delta_{aa'}$. In a modern language, von Neumann's measurement is a conditional expectation onto a maximal Abelian subalgebra of the algebra of all bounded operators acting on the given Hilbert space. As before, $\sum_{a \in M} a\, a^T = \mathbf{1}$. For this type of measurement the Born rule takes a simpler form: $p_a(\rho) = a^T \rho a$. Assuming *ρ* is a pure state this can be simplified further to

$$p_a(\rho) = (a^T \Psi)^2. \quad (2)$$

So, we can see that the probability of the outcome of the measurement will be *a*, if the state is *ρ*, is actually the cosine square of the angle between vectors *a* and *Ψ*, or $p_a(\rho)=\cos^2(a,\Psi)$.

Tsallis' entropy [11] is non-extensive entropy measure proposed in 1988. Given a discrete set of probabilities $\{p_i\}$ and any real number *q*, the Tsallis entropy is defined by the following equation:

$$E_{TS} = \frac{1 - \sum_{k=1}^{K} p_i^q}{q - 1}.$$

Quantum Tsallis' entropy is defined as for *q = r*

$$S_q(\rho) = S^r(\rho) = (1-r)^{-1}(\mathrm{tr}(\rho^r) - 1), r > 0, r \neq 1.$$

## III. Quantum Low Entropy based Associative Reasoning – QLEAR learning

In this section we are going to introduce a Quantum Low Entropy based Associative Reasoning – or QLEAR learning. The QLEAR learning is going to be analyzed in classification problem context. First, in the subsection A, we are going to present a simplified approach in several simple examples where proposed method gives good results. In the subsection B we are going to present a more complex idea that overcomes few drawbacks of the simplified approach.

The approach is based on the idea that classification can be understood as supervised clustering. Clusters analysis divides data into groups (clusters) that are meaningful, useful, or both. If meaningful groups are the goal, then the clusters should capture the natural structure of the data. Classes, or conceptually meaningful groups of objects that share common characteristics, play an important role in how people analyze and describe the world. Human beings (even very small kids) are skilled at dividing objects into groups (clustering) and assigning particular objects to these groups (classification). In the context of understanding data, clusters are potential classes. It is well known that people can quite well generalize the concepts and perform classification based on a few examples. This actually represent the starting point of the method that is going to be analyzed in this paper The basic intention is to "imitate" the principles that are used by humans, and to improve them using some mathematical models that are "not implemented" in the brain, since the brain has the basic function of supporting human survival. In this project, a quantum entropy in the context of the recently proposed probabilistic model proposed in the section II, will be used as a "capturer" (measure, or external index) of the "natural structure" of the data.

### A. Application - a simple classification problem

Here, it will be shown how proposed quantum probabilistic model can be used to solve some classification problems (already presented in [7]). It will be shown how we can solve XOR problem using quantum entropy. We define input vectors as

$$a_1 = \begin{bmatrix} 1 \\ 1 \end{bmatrix},\ a_2 = \begin{bmatrix} -1 \\ -1 \end{bmatrix},\ b_1 = \begin{bmatrix} 1 \\ -1 \end{bmatrix} \text{ and } b_2 = \begin{bmatrix} -1 \\ 1 \end{bmatrix}.$$

Matrix A which describes the system in the plane $z=-1$ (we assume vectors are presented in 3-dimensional system x-y-z) is defined as $A = a_1 * a_1^T + a_2 * a_2^T$, while the system in plane $z=-1$ is defined as $B = b_1 * b_1^T + b_2 * b_2^T$. Then we can calculate the entropy of the individual systems $E_a$ and $E_b$. Overall entropy of the unified system is calculated as $E = E_a + E_b$. Then, for any x and y coordinate, we can calculate how it affects the entropy of the overall system, by adding it to the system A or the system B. We will label it in such a way that addition of the individual point to one of the system creates a minimum change in the overall entropy. The result of such calculation is shown in Fig. 2. From the figure we can see that the problem is successfully solved.

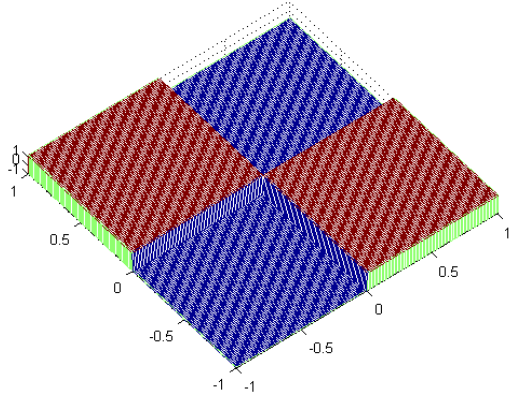

Fig. 2. XOR problem classification

The logical AND problem could be solved in a similar way. However, since we have an unbalanced system it is necessary to define four subclasses and to evaluate the overall entropy for the four possibilities – meaning that any point can potentially belong to any of the four subclasses. The result of classification is shown on Fig. 3.

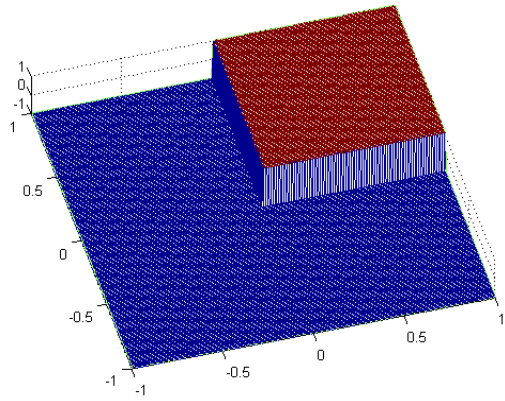

Fig. 3. AND problem classification

Similar method could be used in the famous IRIS classification problem. Taking some points as characteristic representatives (in this case points numbered (20-34) from each group) we can correctly classify the rest of the points. As was the case in the logical AND classification problem, here we have 45 subclasses, and we have to evaluate the entropy for 45 cases – for every point we wont to classify, we can assume that it can belong to any of the 45 subclasses. Although it can look time consuming, we actually have 45 simple independent processes that could be easily realized on parallel hardware, like GPU. For illustration we can use the following figure 6, where we used only attributes 3 and 4 for classification (we can see that this classification is not 100% correct – it is necessary to add attribute 1 too, but we cannot present it graphically).

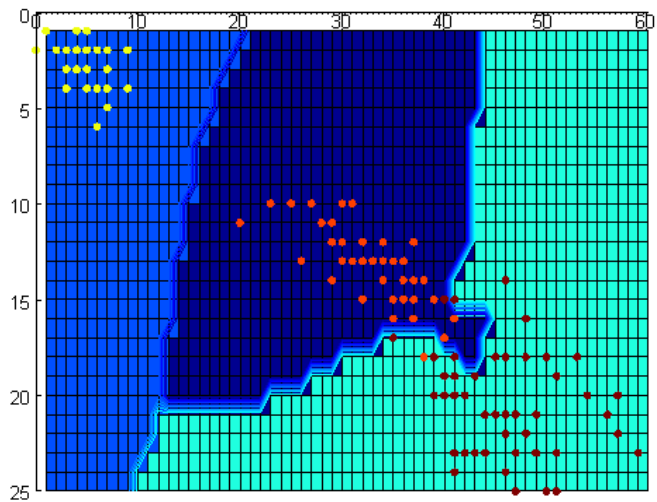

Fig 6. Classification of IRIS data if we use only the data specified by attributes 3 and 4

*B. Application – more complex problems approach*

The approach proposed in the subsection A cannot be easily and successfully implemented on more complex classification problems. There are two obvious drawbacks in that approach – one is, with the exception of XOR problem, representative of the classes were used individually and they were not correlated in any way (they do not 'cooperate') and second is that we are only interested about the similarity with the class representative, but we do not use the opportunity to check what is the level of dissimilarities with the other class representatives, and in a such way we discard potentially useful information.

Here, the category recognition problem represents modification of the approach proposed in [12]. It is done in the framework of measuring similarities to prototype examples of categories, while this approach is quite flexible. While nearest neighbour classifiers are natural in this setting, they suffer from the problem of high variance (in bias-variance decomposition) in the case of limited sampling. Alternatively, one could use some machine learning techniques (like support vector machine), but they usually involve time-consuming optimization, and frequently assume that data could be linearly separated in the hyper plane of proper dimension. We propose a hybrid of these two methods which deals naturally with the multiclass setting, and at the same time has reasonable computational complexity both in training and at run time, and yields excellent results in practice. The basic idea is to find close neighbours to a query sample and then use change in the quantum entropy as a measure of similarity of the newly arrived sample with the representatives of interest. By using quantum entropy we don't make any assumption about linear separability of the data that are going to be classified. At the same time we calculate the dissimilarities with the "most similar" class representatives of the all other groups, and by this, we use most of the information that is available.

The original motivation comes from studies of human perception by Rosch and collaborators [13] who argued that categories are not defined by lists of features, rather by similarity to prototypes. From a machine learning perspective, the most important aspect of this framework is that the emphasis on similarity, rather than on feature spaces, gives us a more flexible framework.

The algorithm of the proposed method could be presented as:

1. Choose a pool of representatives of the $N$ classes represented by sets of vectors $S_1, S_2, \ldots, S_N$ (they do not have necessarily to contain same number of elements).

2. For a given sample, choose the proper value $q$ for the calculation of the Tsallis' entropy, choose a number $N_s$ of the most similar vectors (class representatives) from each pool, $S_i$, that are going to form matrix $\boldsymbol{\rho_s}$ that represent the state of the current class. Also, chose a number $N_{ns}$ (usually smaller than

$N_s$) of the "most similar" class representatives from all other pools $S_j, j \neq i$, that are going to form matrix $\rho_{ns}$ that represent the state of the "complementary" class. Then calculate the entropy of the state $\rho_s$ and state $\rho_{ns}$, and its relative changes $dE_s$ and $dE_{ns}$, by addition of the current sample. Then find for which class the term

$$dE_s - \alpha * dE_{ns}$$

is minimized, and label new sample as a member of that class. The $\alpha$ is a positive real number, usually smaller than 1.

3. Repeat the process for all samples that should be analyzed.

In our case, proper values for $q$, $N_s$, $N_{ns}$ and $\alpha$ were selected from 2-fold cross validation (CV) process that is implemented on training data represented by the union of sets $S_i$.

## IV. Experiments and Results

In this section we are going to present results for several classification problem. Classification results were based on the method proposed in Section III. In all examples the maximum size of the pool of the representatives was the half of the data from that class.

Sets that were analyzed (taken from [14]), together with classification results were presented in the following Table I. We will stress that results are improved by application of several techniques, that could be understood as engineering of the proposed theoretical model, and which are not going to be analyzed here.

TABLE I – RESULTS OF CLASSIFICATION FOR SEVERAL PUBLICALLY AVAILABLE DATA SETS

|  | Pool size - max | $N_s$ | $N_{ns}$ | $q$ | Error[%] |
|---|---|---|---|---|---|
| Appendix | 35 | 7 | 4 | 0.03 | 0 |
| Australian Credit | 195 | 7 | 5 | 0.11 | 0 |
| Banana | 1500 | 26 | 1 | 1.78 | 3.2 |
| Contraceptive | 320 | 29 | 1 | 1.5 | 9.7 |
| Glass | 38 | 5 | 2 | 1.22 | 2.8 |
| German Credit | 265 | 25 | 1 | 0.5 | 0 |
| Iris | 21 | 14 | 1 | 2 | 0 |
| Parkinson | 40 | 14 | 5 | 0.03 | 0.003 |
| Pima | 230 | 13 | 1 | 0.95 | 0 |
| Wine | 25 | 9 | 6 | 0.03 | 0 |

From the table we can see, that results of the classification could be considered satisfactory. Here must be stressed that values for most of the parameters are probably not optimal, since the goal of this research was to explore the potential of the proposed paradigm, and not to find optimal values for the parameters. For instance, it is possible to achieve correct classification for Iris data set for $N_s = 8$, with proper selection of the representative vectors. However, minimization of the number of representative vectors is out of the scope of this paper.

## V. Conclusion

In this paper new learning paradigm, called QLEAR learning was introduced and applied on classification problem where data is represented by vectors. Based on presented results we can conclude that proposed paradigm has some potential in application where classification of vectors is required.

Since the proposed paradigm is generic in nature, it is clear that full potential of the proposed learning method requires further analyses. There are several aspects of the proposed method that could be improved and some of them are quite obvious. For instance, it could be possible to create an on-line adaptive method that search for optimal number of the class representatives that is optimized after every sample (or every few samples). Also, it could be analyzed is the Euclidean distance the optimal choice of the similarity measure or is the Tsallis' entropy the optimal choice of quantum entropy measure.

Extension of the proposed learning method to the data represented with matrices and higher-order multiway arrays could be also analyzed and it would represent the direction of future research.